\documentclass[conference]{IEEEtran}
\usepackage{times}

\usepackage[numbers]{natbib}
\usepackage{multicol}
\usepackage[bookmarks=true]{hyperref}

\usepackage{graphicx}
\usepackage{tcolorbox}
\usepackage{booktabs}
\usepackage{amssymb}
\usepackage{amsmath}
\AtEndPreamble{
    \usepackage[capitalize]{cleveref}
    \crefname{section}{Sec.}{Secs.}
    \Crefname{section}{Section}{Sections}
    \Crefname{table}{Table}{Tables}
    \crefname{table}{Tab.}{Tabs.}
}

\usepackage{caption}  %

\begin{document}

\title{\fontsize{23.5pt}{20pt}\selectfont GeneralVLA: Generalizable Vision–Language–Action Models with Knowledge-Guided Trajectory Planning}

\author{
    \textbf{Guoqing Ma}$^{1*}$ \quad
    \textbf{Siheng Wang}$^{2*}$ \quad
    \textbf{Zeyu Zhang}$^{2*\dag}$ \quad
    \textbf{Shan Yu}$^{1}$ \quad
    \textbf{Hao Tang}$^{2\ddag}$ \vspace{0.1em}\\
    $^{1}$CASIA \quad
    $^{2}$Peking University\\
    \small $^*$Equal contribution. $^\dag$Project lead.
    $^\ddag$Corresponding author: bjdxtanghao@gmail.com.
}

\maketitle

\begin{abstract}
Large foundation models have shown strong open-world generalization to complex problems in vision and language, but similar levels of generalization have yet to be achieved in robotics.
One fundamental challenge is that the models exhibit limited zero-shot capability, which hampers their ability to generalize effectively to unseen scenarios.
In this work, we propose GeneralVLA (Generalizable Vision–Language–Action Models with Knowledge-Guided Trajectory Planning), a hierarchical vision-language-action (VLA) model that can be more effective in utilizing the generalization of foundation models, enabling zero-shot manipulation and automatically generating data for robotics.
In particular, we study a class of hierarchical VLA model where the high-level ASM (Affordance Segmentation Module) is finetuned to perceive image keypoint affordances of the scene; 
the mid-level 3DAgent carries out task understanding, skill knowledge, and trajectory planning to produce a 3D path indicating the desired robot end-effector trajectory. 
The intermediate 3D path prediction is then served as guidance to the low-level, 3D-aware control policy capable of precise manipulation.
Compared to alternative approaches, our method requires no real-world robotic data collection or human demonstration, making it much more scalable to diverse tasks and viewpoints.
Empirically, GeneralVLA successfully generates trajectories for 14 tasks, significantly outperforming state-of-the-art methods such as VoxPoser. The generated demonstrations can train more robust behavior cloning policies than training with human demonstrations or from data generated by VoxPoser, Scaling-up, and Code-As-Policies.
We believe GeneralVLA can be the scalable method for both generating data for robotics and solving novel tasks in a zero-shot setting.
Code: \url{https://github.com/AIGeeksGroup/GeneralVLA}.
Website: \url{https://aigeeksgroup.github.io/GeneralVLA}. 
\end{abstract}

\IEEEpeerreviewmaketitle

\section{Introduction}
\label{sec:intro}

Developing robot manipulation policies that generalize in a zero-shot manner remains a long-standing challenge. 
Foundation models can be viewed as compressed world models, through which humans distill and transmit accumulated knowledge and experience about the physical world. 
Motivated by the strong generalization capabilities exhibited by large vision-language models (VLMs), recent work has explored whether this paradigm can be directly extended to robot manipulation. A line of prior studies~\cite{DBLP:conf/corl/ZitkovichYXXXXW23, DBLP:conf/corl/KimPKXB0RFSVKBT24, DBLP:journals/corr/abs-2410-24164} proposes open-world vision-language-action models (VLAs) by finetuning pretrained VLMs to output robot actions in an end-to-end manner. 
We refer to these approaches as \textit{monolithic VLA models}. 
Such models critically depend on large-scale robotics datasets that couple on-robot sensory observations, including visual inputs and proprioceptive states, with corresponding action trajectories.

However, current monolithic VLA models are incapable of achieving off-domain zero-shot generalization at a level comparable to that of VLMs and LLMs in their research domains~\cite{intelligence2504pi0, zhao2025cot, yang2025qwen3}.
Moreover, even after fine-tuning, VLMs tend to underperform compared to other models. Because they cannot provide fine-grained coordinates, unlike task-specific models(eg,. SAM~\citep{sam}, YOLO~\citep{yolov11}, Deformable-DETR~\citep{deformable}.), although they possess an understanding capability.
In addition, planning long-horizon manipulation trajectories in 3D space without demonstrations remains challenging due to the gap between high-level semantic reasoning and continuous geometric constraints~\cite{saycan, code}, as well as the inability of existing systems to accumulate and reuse experience across tasks~\cite{inner}. Besides, monolithic VLA models have yet to demonstrate emergent generalization comparable to VLMs and LLMs in other domains of study and are are constrained by their inference frequency to achieve dexterous and dynamic manipulation tasks~\cite{DBLP:conf/corl/ZitkovichYXXXXW23, DBLP:conf/corl/KimPKXB0RFSVKBT24}.

\begin{figure}[t]
  \centering
  \includegraphics[width=1. \linewidth]{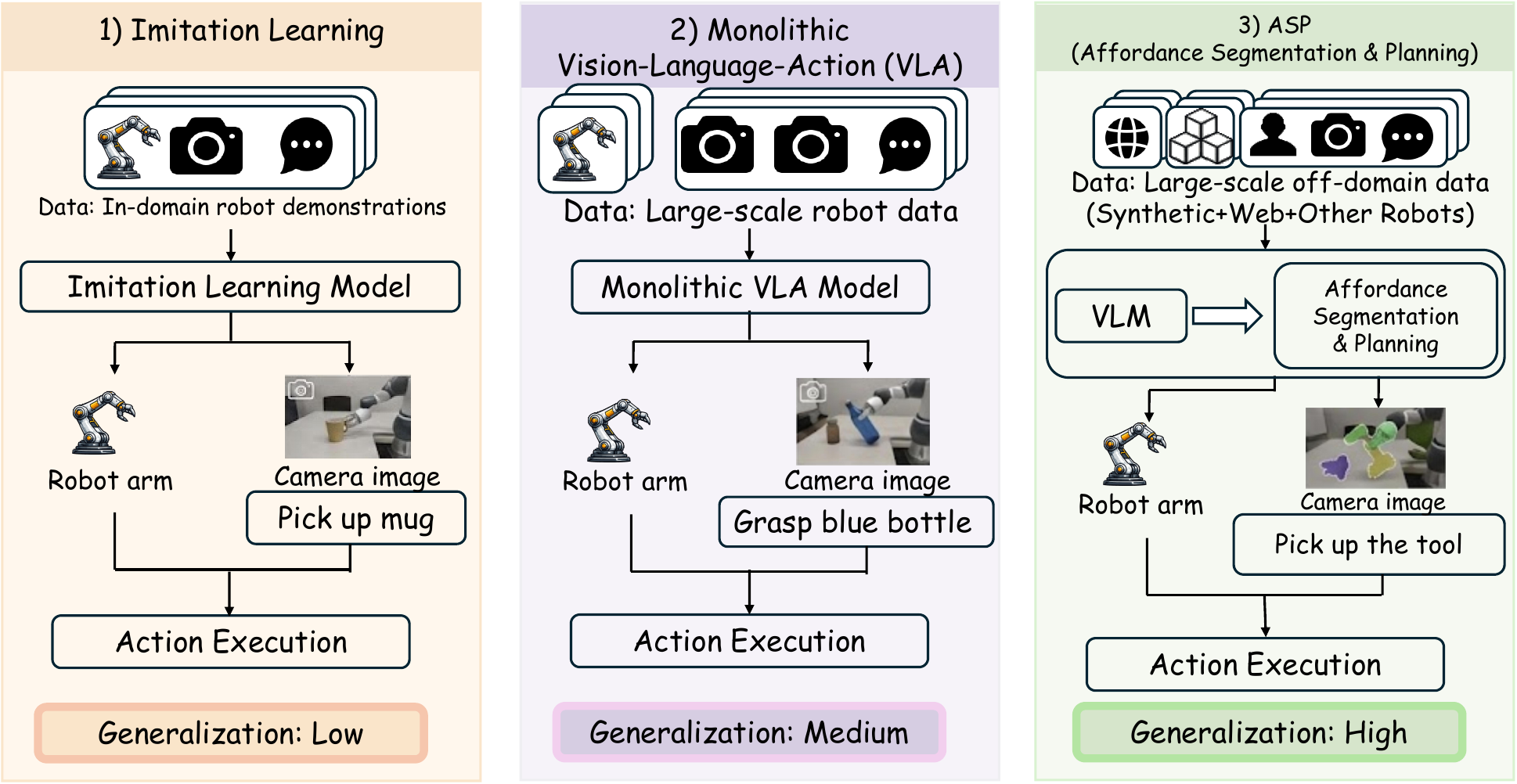}
  \vspace{-20pt} 
  \caption{\textbf{Overview of GeneralVLA, VLAs and earlier imitation learning methods.} GeneralVLA’s hierarchical design results in better generalization.
  It enables 3D trajectory planning framework that fully exploits the prior knowledge of foundation models.
  } %
  \label{fig:intro}
    \vspace{-20pt} 
\end{figure}

To address the above challenges, we propose a hierarchical architecture for VLA, GeneralVLA (Generalizable Vision–Language–Action Models with Knowledge-Guided Trajectory Planning), 
which can complete robotic manipulation tasks while generating rich robotic data in a zero-shot manner.
Specifically, the high-level ASM (Affordance Segmentation Module) leverages a highly generalized vision-language-model (VLM) and the segment anything model (SAM) for task understanding and identifying the affordance positions of key objects. 
In addition, By converting these affordances into 3D representations via a depth map, the mid-level Knowledge-Guided Trajectory Planning carries out task understanding, skill knowledge, and trajectory planning to generate a 3D path that outlines the desired trajectory for the robot's end-effector.
Furthermore, in order to fully utilize the experience of multitasking execution, we have designed a knowledge bank that enables it to summarize and store common skills.
We term the combination of ASM and Knowledge-Guided Trajectory Planning the Hierarchical World Model.
Both ASM and Knowledge-Guided Trajectory Planning are foundation models infused with massive world priors and are dedicated to trajectory planning like classic world models.
The intermediate 3D path planning then serves as guidance for the low-level, 3D-aware control policy capable of precise manipulation.
The low-level policy is executed by estimating precise grasp poses. We design HGM to sharpen grasping accuracy.

The hierarchical architecture has multiple benefits.
ASM integrates the generalization capabilities of VLM and SAM, enabling more precise multi-object affordance localization.
The hierarchical design presented in GeneralVLA also offers additional advantages through the decoupling of ASM finetuning, LLM planning and low-level action prediction. 
Specifically, while the higher-level ASM is predicting semantically meaningful 2D affordance point from RGB camera inputs, 
it does not need to sacrifice visual understanding to preserve long horizon planning capability.  
The mid-level Knowledge-Guided Trajectory Planning is used for long horizon planning. By leveraging the strong textual generalization, it enables spatial reasoning, pose estimation, and trajectory planning.
The lower-level policy models can additionally operate from rich 3D and proprioceptive information. 
In doing so, GeneralVLA inherits the general visual perception ability of VLM and SAM, semantic reasoning benefits of LLMs and the 3D spatial awareness benefits of 3D policy models~\cite{DBLP:conf/rss/0001B0GCF24, DBLP:conf/corl/KeGF24}.

Our contributions are as follows:

\begin{itemize}
\item We propose a zero-shot 3D trajectory planning framework that solves robotic manipulation tasks while generating rich robotic data. This paper achieves it via a hierarchical VLA that fully exploits the prior knowledge of foundation models.

\item We propose an ASM (Affordance Segmentation Module) that leverages VLM and SAM foundations with iterative refinement for accurate affordance segmentation.
Additionally, a knowledge bank is introduced in Knowledge-Guided Trajectory Planning to capture cross-task skills, thereby improving success rates of manipulation tasks.

\item Experiments show that our method achieves high zero-shot accuracy on diverse manipulation tasks, outperforming state-of-the-art methods such as VoxPoser. Moreover, benchmarking experiments demonstrate that the data generated by GeneralVLA is high-quality and scalable.

\end{itemize}

\section{Related Work}

\textbf{LLMs and VLMs for robotics.} Early attempts in leveraging LLMs or VLMs for robotics are through pretrained language~\cite{DBLP:conf/icra/SinghBMGXTFTG23} and visual~\cite{DBLP:conf/icml/ShahK21, DBLP:conf/icml/ParisiRP022, DBLP:conf/corl/NairRKF022, DBLP:conf/iclr/MaSJBK023} representations. 
However, these are insufficient for complex semantic reasoning and generalization to the new scene in a zero-shot manner. 
Recent research has focused on directly leveraging open world reasoning and generalization capability of LLMs and VLMs, by prompting or fine-tuning them to, e.g., generate plans~\cite{DBLP:conf/icra/LiangHXXHIFZ23, DBLP:conf/corl/DuanYPWEFK24, DBLP:conf/corl/HuangXXCLFZTMCS22, DBLP:journals/arobots/LinAMPB23, DBLP:conf/icra/SinghBMGXTFTG23, DBLP:conf/corl/IchterBCFHHHIIJ22} or construct value~\cite{DBLP:conf/corl/HuangWZL0023} and reward functions~\cite{DBLP:conf/iclr/KwonXBS23, DBLP:conf/nips/SontakkeZAPBSFI23, DBLP:conf/corl/0003GFKLACEHHIX23, DBLP:conf/iclr/MaLWHBJZFA24, DBLP:conf/icml/WangSZXBHE24}.
However, LLM planners are limited to high-level macro planning, with success rates heavily dependent on the control primitives~\cite{DBLP:conf/icra/LiangHXXHIFZ23}.
Besides, VLM-based methods (e.g., VoxPoser~\cite{DBLP:conf/corl/HuangWZL0023}) can only produce simple value functions and struggle with complex, long-horizon tasks.
Our work simultaneously leverages VLM and LLM, enabling long-horizon planning in 3D space while incorporating precise motion control.

\begin{figure*}
  \centering
  \vspace{-15pt} 
  \includegraphics[width=.9\textwidth]{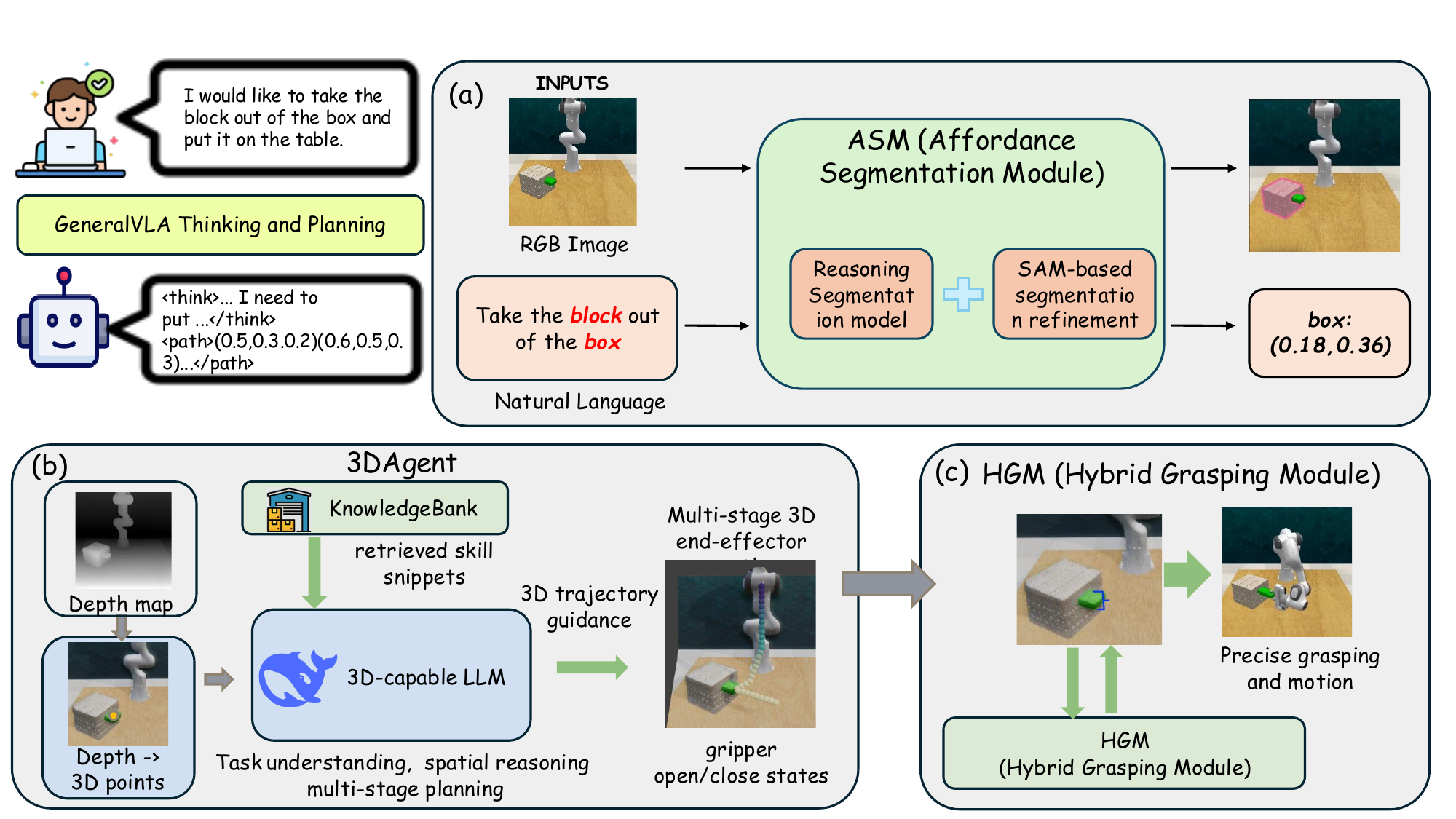}
  \vspace{-5pt} 
  \caption{\textbf{Inference workflow of of GeneralVLA.} (a) The high-level ASM is called to generate the 2D points and corresponding semantic information. %
  (b) The mid-level Knowledge-Guided Trajectory Planning carries out task understanding, 3D reasoning and planning to produce a 3D path indicating the desired robot end-effector trajectory. 
  (c) The intermediate 3D path prediction is then served as guidance to the low-level, 3D-aware control policy enhanced by HGM for precise manipulation.
 } %
  \label{fig:pipline}
    \vspace{-15pt} 
\end{figure*}

\textbf{Monolithic VLA models as language-conditioned robot policies.} Monolithic VLA models have been proposed to produce robot actions given task descriptions and image observations directly~\cite{DBLP:conf/rss/BrohanBCCDFGHHH23, DBLP:journals/corr/abs-2210-03094, DBLP:conf/corl/ZitkovichYXXXXW23, DBLP:conf/rss/GhoshWPBMDHK0LT24, DBLP:conf/corl/KimPKXB0RFSVKBT24, DBLP:conf/corl/RadosavovicSFGD23}. Monolithic VLA models are often constructed from VLMs~\cite{DBLP:conf/nips/LiuLWL23a, DBLP:journals/corr/abs-2502-13923, DBLP:conf/icml/DriessXSLCIWTVY23, DBLP:conf/cvpr/LinYP0SH24}, and are trained on large-scale on-robot data~\cite{DBLP:conf/rss/BrohanBCCDFGHHH23, DBLP:conf/icra/ONeillRMGPLPGMJ24, DBLP:conf/rss/KhazatskyP0BDKN24} to predict actions as text or special tokens. However, due to the lack of coverage in existing robotics datasets, they must be finetuned in-domain on expensive on-robot data. Their action frequency is also constrained by inference frequency, limiting their capability to achieve dexterous and dynamic tasks. The most relevant monolithic VLA model to our work is LLARVA~\cite{DBLP:conf/corl/NiuSBQBSDH24}, which predicts end-effector trajectories in addition to robot actions. However, LLARVA only uses trajectory prediction as an auxiliary task to improve the action prediction of a monolithic VLA model. 
In contrast, our work takes a hierarchical approach that can preserve semantic reasoning when fine-tuning the ASM, while avoiding any trade-off between stronger reasoning and visual understanding.
It enables us to fully exploit the prior knowledge of foundation models, leverage abundant and inexpensive off-domain data, and achieve zero-shot task completion.

\textbf{VLMs for predicting intermediate representations.} 
Point-based predictions: A common intermediate prediction interface has been keypoint affordances~\cite{DBLP:conf/corl/StoneXLGLVWKZXF23, DBLP:conf/corl/SundaresanBSB23, DBLP:conf/icml/NasirianyX0XL0X24, DBLP:conf/corl/YuanDBPKMMF24, DBLP:conf/corl/KuangYGMDG0024}. Keypoint affordances can be obtained through the use of open-vocabulary detectors~\cite{DBLP:conf/eccv/MindererGSNWDMA22}, iterative prompting of VLMs~\cite{DBLP:conf/icml/NasirianyX0XL0X24}, or fine-tuning detectors to identify certain parts of an object semantically~\cite{DBLP:conf/corl/SundaresanBSB23}. 
As opposed to these, our work 
plans 3D paths in 3D space, leveraging the powerful reasoning capabilities of LLM.

Trajectory-based predictions: The idea of using trajectory-based task specifications to condition low-level policies was proposed in RT-trajectory~\cite{DBLP:conf/iclr/GuKW0AR0FGXSX0H24}, largely from the perspective of flexible task specification. 
Similarly, the emergence of track-any-point (TAP) models~\cite{DBLP:conf/iccv/DoerschYVG0ACZ23, DBLP:conf/iccv/WangCCLHHS23} has enabled policies conditioned on object trajectories~\cite{DBLP:conf/corl/YuanWZG24, DBLP:conf/corl/XuXXCWVS24, DBLP:journals/corr/abs-2405-01527} or points sampled from a fixed grid in the image~\cite{DBLP:conf/rss/WenLS0D0A24}. 
Our work performs trajectory planning in 3D space and enables obstacle avoidance and global planning, achieving zero-shot completion. 

\textbf{Leveraging simulation data for training robot policies.} There has been extensive work on leveraging simulation for robot learning. Simulation data is popular in reinforcement learning (RL), as RL on real robotic systems is often impractical due to high sample complexity and safety concerns~\cite{DBLP:journals/scirobotics/HwangboWK020, DBLP:conf/icra/HandaAMPSLMWZSN23, DBLP:conf/rss/VillasevilSLC0024}. Recently, simulation has been also exploited to directly generate~\cite{DBLP:conf/corl/FishmanMEPBF22} or bootstrap~\cite{DBLP:conf/corl/MandlekarNWANFZ23} large-scale datasets for imitation learning, to reduce the amount of expensive robot teleoperation data needed. 
Our work harnesses the inherent capabilities of VLM, SAM and LLM as a world model for planning, thereby generating simulation data. This simulation data is massive, low-cost, and of high quality with a high success rate.

\section{The Proposed GeneralVLA}

\subsection{GeneralVLA’s Framework}
In this work, we examine how VLA models can leverage 
prior knowledge of foundation models and demonstrate cross-domain transfer capabilities, as opposed to relying purely on expensive observation-language action data collected on a robot. 
GeneralVLA is a hierarchical VLA model designed for this purpose, exhibiting generalizable and robust manipulation. 
It consists of three interconnected models (\cref{fig:pipline}): first, a higher-level ASM produces point affordance guidance (detailed in~\cref{Affordance Segmentation Module}); second, a mid-level 3DAgent serves as a general knowledge world model to produce 3D path guidance (detailed in~\cref{Knowledge-Guided Trajectory Planning}); and third, a low-level policy that produces actions conditioned on 3D paths (detailed in~\cref{Path Guided Low-level policy learning}).

\begin{figure*}
  \centering
  \includegraphics[width=.9\textwidth]{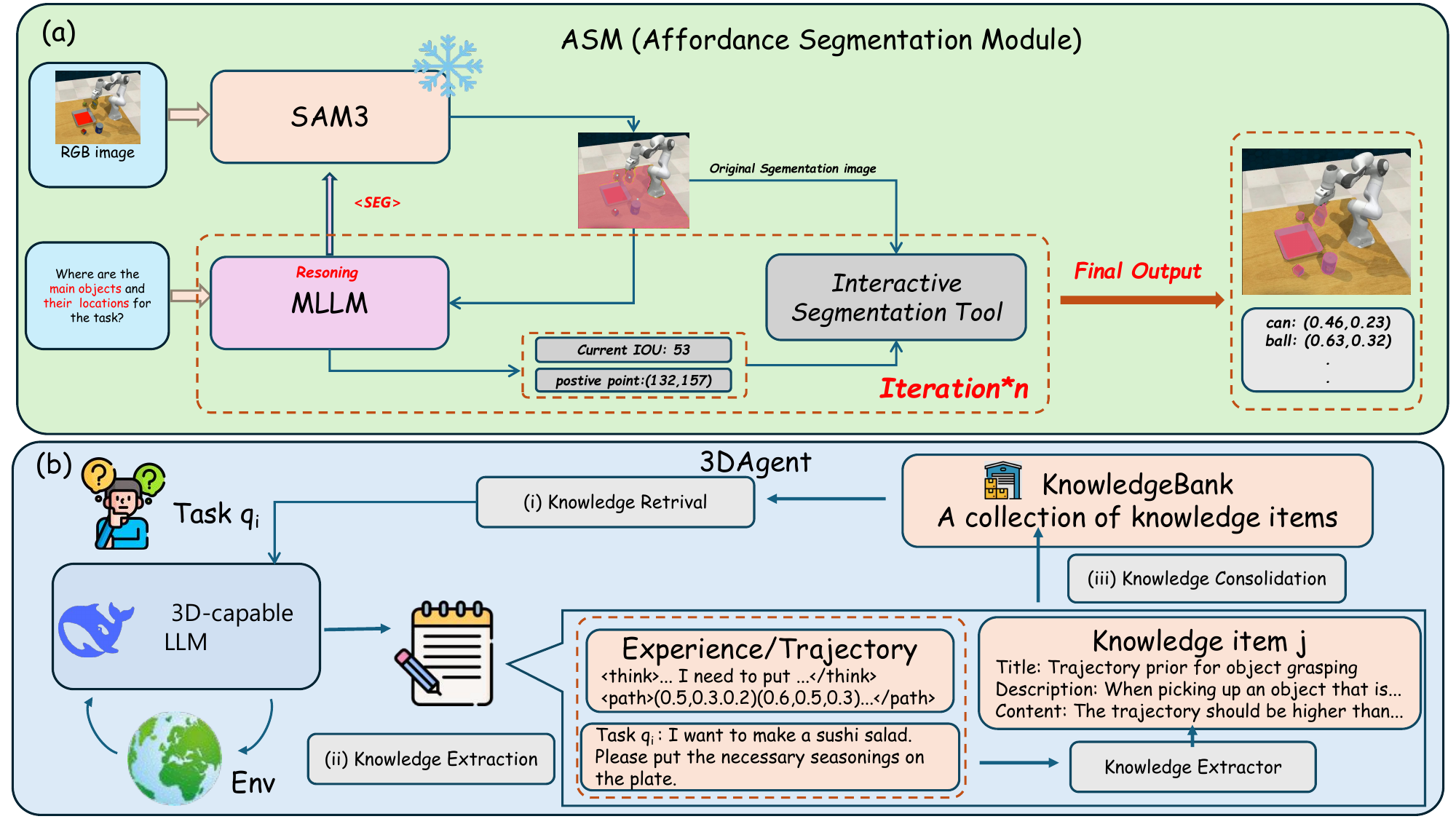}
  \caption{
  \textbf{Detailed framework of ASM and 3DAgent. } (a) Given the input image and task text as query, the multimodal LLM (e.g., LLaVA~\citep{llavanext}) generates text output. The last-layer embedding for the \texttt{<SEG>} token is then decoded into the segmentation mask via the decoder. We use LoRA~\citep{LoRA} for efficient fine-tuning. The choice of vision backbone can be flexible (e.g., SAM3~\citep{sam3}). 
  }
  \label{fig:detailed_pipeline}
    \vspace{-15pt} 
\end{figure*}

\subsection{Affordance Segmentation Module}
\label{Affordance Segmentation Module}

ASM is used to perceive the current scene, understand the affordance of objects and obstacles, and mark them in the form of points (\cref{fig:pipline}a). This facilitates understanding and planning by the following 3DAgent. Traditional VLM models already possess certain capabilities for recognizing the 2D positions of objects~\cite{DBLP:conf/corl/YuanDBPKMMF24, team2025gemini}, but their ability to locate the precise positions of objects and their affordances is still relatively weak, failing to meet the demands of high-precision operations. To address this, 
we incorporate SAM's segmentation priors and design a refinement process. To achieve finer-grained segmentation results (e.g., identifying graspable regions of a robotic arm in an image), we first leverage the reasoning capabilities of a multimodal large language model (MLLM) to guide the segmentation process of SAM~\cite{lisa}, as illustrated in Fig~\ref{fig:detailed_pipeline}. When the LLM is required to produce a binary segmentation mask, the generated output sequence includes a special \texttt{<SEG>} token. We extract the hidden representation $\tilde{h}_{\text{seg}}$ from the last layer of the LLM corresponding to this \texttt{<SEG>} token, and project it through an MLP to obtain the segmentation embedding $h_{\text{seg}}$. The overall process can be summarized as: 
\begin{equation}
\begin{aligned}
h_{\text{seg}} &= \gamma(\tilde{h}_{\text{seg}}), \\
f & = \mathcal{F}_{\text{enc}}(x_{\text{img}}), \\
\hat{M} & = \mathcal{F}_{\text{dec}}(h_{\text{seg}}, f).
\end{aligned}
\end{equation}
where the $\hat{M}$ denotes the final segmentation mask, and the \text{enc} and \text{dec} refer to the SAM's Encoder and Decoder. However, a single-pass segmentation may suffer from over-segmentation or under-segmentation, which degrades the accuracy of the extracted 2D coordinates. Such errors are further propagated to the reconstructed 3D geometry, resulting in discrepancies between the planned and actual 3D trajectories, and ultimately leading to execution failures of the robotic manipulator. To address this issue, following SegAgent~\cite{segagent}, we propose an iterative segmentation refinement mechanism. Specifically, after the Interaction Segmentation Tool produces an initial segmentation, a multimodal large language model (MLLM) evaluates the result and provides two types of feedback points: positive points, indicating regions that are correctly segmented, and negative points, indicating regions with segmentation errors. These points are then used to guide the Interaction Segmentation Tool in refining its prediction in the next iteration. This interaction process is repeated for up to $n$ iterations. When no negative points are generated, the segmentation is considered sufficiently accurate, and the corresponding result is taken as the final output.

\begin{figure*}
  \centering
  \includegraphics[width=.95\textwidth]{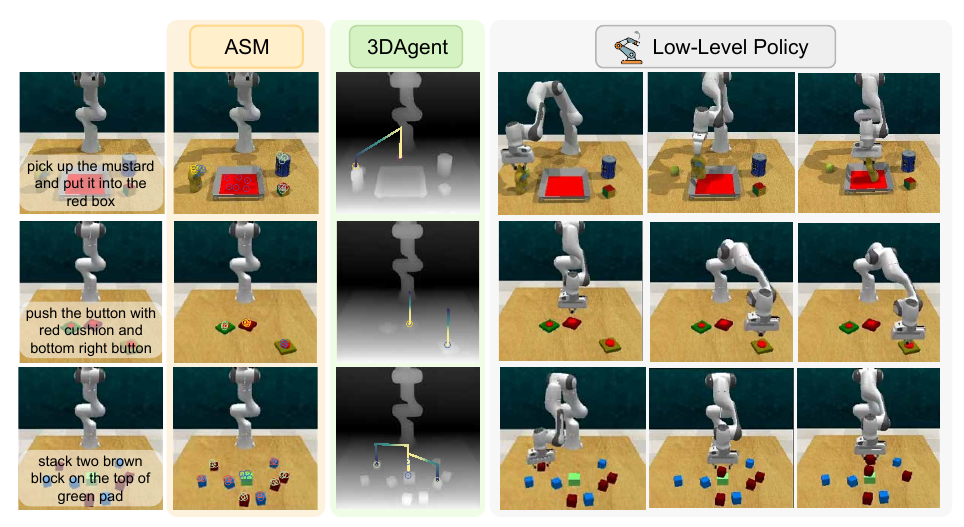}
  \vspace{-10pt} 
  \caption{\textbf{Example GeneralVLA rollouts} demonstrate its strong performance in multi-object, multi-stage scenes, achieved by leveraging ASM’s segmentation capability, 3DAgent’s spatial reasoning ability, and the robust execution of the low-level 3D policy.} %
  \label{fig:examplerollout}
    \vspace{-10pt} 
\end{figure*}

\subsection{Knowledge-Guided Trajectory Planning} %
\label{Knowledge-Guided Trajectory Planning}

Due to the insufficient capability of existing visual foundation models in understanding and planning in 3D scenes, we leverage the textual information of 3D points and utilize the powerful text generalization ability of LLMs to solve the 3D path planning problem. 
To fully utilize the experience of multitasking execution, we have designed a knowledge bank that enables it to summarize and store common skills.
We refer to this process as 3DAgent (\cref{fig:pipline}b).
The information from ASM includes 2D point information and object semantics. We use depth information to project the 2D points into the depth field to obtain 3D information, thereby acquiring 3D point information along with object semantics. The task instruction, 3D point information, and object semantics are then input into the LLM. 

The LLM can understand the task, comprehend the scene, and plan the trajectory of the robotic arm. The trajectory includes states for the gripper being closed and open. A motion trajectory can consist of multiple stages, enabling long-horizon planning and avoiding the discrete sub-tasks issues in traditional planning method, as well as the problems arising from stitching discrete sub-tasks~\cite{DBLP:conf/icra/LiangHXXHIFZ23}. Additionally, the motion trajectory can satisfy obstacle avoidance requirements.
In our experiments, when the number of points per object exceeds 3, the LLM can effectively understand the spatial pose of the object. %

3DAgent is equipped with KnowledgeBank. The integration proceeds in three steps: (i) knowledge retrieval, (ii) knowledge construction,
and (iii) knowledge consolidation, as shown in~\cref{fig:pipline}b. 
During knowledge retrieval, the agent queries
KnowledgeBank with the current query context to identify the top-k relevant experiences and their
corresponding knowledge items using embedding-based similarity search. 
Retrieved items are injected
into the agent’s system instruction, ensuring that the manipulation is grounded in useful past
experiences. 
When the current query task is completed, we will perform knowledge construction to
extract new knowledge items. 

The first step is to obtain proxy signals for the correctness of completed
trajectories: we adopt an LLM-as-a-judge~\cite{gu2024survey} to label outcomes as success or failure, given the query and trajectory, without access to any ground-truth. 
Based on these signals, we apply
different extraction strategies: successful experiences contribute validated manipulation strategies, while failed ones supply counterfactual signals and pitfalls that help sharpen guardrails. 
In practice, we extract multiple knowledge items for each trajectory/experience (\cref{fig:detailed_pipeline}b).
Finally, knowledge consolidation incorporates these items into KnowledgeBank with a simple addition operation, maintaining an evolving repository of knowledge items. 
Together, these steps form a closed-loop process: the agent leverages past experiences, constructs new knowledge from current tasks, and continually updates its knowledge, enabling sustained evolution in test-time learning scenarios.

\subsection{Path Guided Low-level policy} %
\label{Path Guided Low-level policy learning}
The planned 3D Path is a macroscopic coarse trajectory. When grasping an object at a grasp point in 3D space, precise grasp pose estimation is required to ensure robust grasping of different objects. To enhance the precision of robotic arm operations, we designed a multimodal hybrid grasping module (HGM). 

The 3D point information can locate the 3D spatial range of the object. After determining the 3D spatial range, RGB color and depth are fused, and point cloud data is obtained using the inverse projection method. Once the object point cloud data is determined, we use the grasp prediction model~\cite{DBLP:conf/corl/YuanMMF23} to estimate the grasp pose.
Additionally, since graspnet provides multiple candidate grasp points, we filter out those that would cause collisions and select the grasp pose whose grasp center is closest to the object's center point.
This approach allows us to obtain the optimal task-specific grasp pose, %
all leveraging the capabilities of the foundation model.
Due to cropping the point cloud using a 3D spatial range, the speed of grasp pose estimation is improved.
After the action is generated, a motion planner can be used to move the robot to the desired pose, as detailed in~\cref{fig:pipline}c.

\begin{table*}[t]
\setlength{\tabcolsep}{4pt}
\caption{\textbf{Task-averaged success rate \% for zero-shot evaluation.} GeneralVLA outperformed other baselines in 10 out of 14 simulation tasks from RLBench~\cite{DBLP:journals/ral/JamesMAD20}. 
w/o PA indicates that only the pre-tuned VLM is used for localization without ASM.
Each task was evaluated over 3 seeds to obtain the task-averaged success rate and standard deviations.}
\centering
\vspace{-5pt} 
    \resizebox{1.\textwidth}{!}{
    \begin{tabular}{cccc cccc} %
      \toprule  %
      \enspace \textbf{Method}   & Put\_block & Play\_jenga & Open\_jar &Close\_box & Open\_box & Pickup\_cup &Push\_block  \\
      \midrule
      VoxPoser~\cite{DBLP:conf/corl/HuangWZL0023} &   70.70±2.31 & 0.00±0.00  & 0.00±0.00 & 0.00±0.00 & 0.00±0.00 & 26.70±14.00  & \textbf{25.33}±8.33 
\\
      CAP~\cite{DBLP:conf/icra/LiangHXXHIFZ23} &  84.00±16.00 & 0.00±0.00 & 0.00±0.00 & 0.00±0.00 & 0.00±0.00 &  14.67±4.62 & 8.00±4.00
\\
      Scaling-up~\cite{DBLP:conf/corl/HaFS23} &  77.33±6.11  & 0.00±0.00 & 78.67±11.55 & 0.00±0.00 & 0.00±0.00 & 9.33±2.26  & 5.33±6.11
\\
      GeneralVLA (Ours) &  \textbf{93.33}±3.06  & \textbf{84.67}±11.02 & \textbf{84.00}±3.46 & \textbf{52.00}±11.14 & \textbf{35.33}±12.86 & \textbf{88.67}±1.15 & 22.67±14.05
\\
      GeneralVLA w/o PA &  72.00±7.21  & 60.67±9.45 & 67.33±12.06 & 28.67±7.57 & 8.67±4.16 & 74.00±10.58 & 12.00±8.72
\\
      \bottomrule %
      \toprule
        \enspace \textbf{Method}   & Take\_umbrella & Sort\_mustard & Open\_wine & Lamp\_on & Put\_knife & Pick\_\&\_lift & Insert\_block  \\
      \midrule
      VoxPoser~\cite{DBLP:conf/corl/HuangWZL0023} &   33.33±8.33 & \textbf{96.00}±6.93  & 8.00±4.00 & 57.30±12.22 &  \textbf{92.00}±4.00 & 96.00±0.00  & 0.00±0.00
\\
      CAP~\cite{DBLP:conf/icra/LiangHXXHIFZ23} &  4.00±4.00 & 0.00±0.00 & 0.00±0.00 & 64.00±6.93 & 14.67±8.33 & \textbf{ 100.00}±0.00  & 0.00±0.00
\\
      Scaling-up~\cite{DBLP:conf/corl/HaFS23} & 6.67±2.31  & 41.33±12.86 & 33.33±20.13 & 60.00±8.00 & 24.00±0.00 & \textbf{100.00}±0.00  & 0.00±0.00
\\
      GeneralVLA (Ours) &  \textbf{67.33}±14.05  & 76.00±15.62 & \textbf{47.66}±11.37 & \textbf{74.67}±11.37 & 60.67±19.22 & 92.00±6.00  & \textbf{32.67}±6.43
\\
      GeneralVLA w/o PA &  46.00±12.00  & 56.00±8.72 & 22.00±11.14 & 57.33±15.53 & 44.00±4.00 & 64.67±12.86 & 13.33±4.16
\\
      \bottomrule %
    \end{tabular}
    }
    \label{table:zero-shot}
    \vspace{-10pt} 
\end{table*}

\section{Experiments}

Our experiments are designed to address two questions: 1) Can GeneralVLA accurately solve a diverse set of tasks in a zero-shot manner? 2) Can data generated from GeneralVLA be used to train a robust policy?

\noindent \textbf{Implementation details.} 
We use our ASM to detect and extract object information. 
To estimate object positions more accurately and enable the language model to estimate the object's 3D pose, each object is sampled with at least three points.
We use Deepseek R1 for reasoning and planning on pure text-based 3D points.
To ensure zero-shot execution within a reasonable budget, we limit the number of 3D points in the 3D path to 20. This also ensures the stability of the planning process and avoids excessively long planning.
Full prompts are included in the Appendix. 
To theoretically validate the feasibility of the GeneralVLA framework without overly increasing system complexity, we used only the front viewpoint, which typically offers the best perspective.
For better reasoning by the VLM, we use a resolution of 256 × 256.

\subsection{Zero-shot Performance in Simulation}
\label{Zero-shot Performance in Simulation}

We empirically study the zero-shot capability of GeneralVLA in solving 14 diverse tasks in simulation, covering a wide range of task configurations and action primitives for both prehensile and non-prehensile tasks. Our simulation experiments are reported to ensure reproducibility and provide a benchmark for future methods.

\noindent \textbf{Environment and tasks.} The simulation setup involves a Franka Panda robot with a parallel gripper, using CoppeliaSim and PyRep for interfacing. Four RGB-D cameras capture input observations in front of a tabletop environment. RLBench~\cite{DBLP:journals/ral/JamesMAD20} is used as the task benchmark, with 14 sampled tasks that cover various object category, object position variations and task horizons. The robot’s actions are represented as waypoints, with trajectories computed and executed via a motion planner~\cite{DBLP:journals/ram/SucanMK12}.

\noindent \textbf{Baselines.} We compare against three state-of-the-art zero-shot data generation approaches: Code-as-Policies (CAP)~\cite{DBLP:conf/icra/LiangHXXHIFZ23}, Scaling-up-Distilling-Down (Scaling-up)~\cite{DBLP:conf/corl/HaFS23} and VoxPoser~\cite{DBLP:conf/corl/HuangWZL0023}. CAP uses language models to generate programs that call hand-crafted primitive actions, while VoxPoser predicts waypoints via a 3D voxel map of value functions. Scaling-up leverages an LLM with 6 DoF exploration primitives to generate robotic data for policy distillation. We provided CAP and Scaling-up with ground truth simulation states and object models, and VoxPoser with segmented object point clouds, which inherently disadvantages GeneralVLA in comparison.

\noindent \textbf{Results: GeneralVLA can generate successful trajectories for all 14 tasks, while Scaling-up, VoxPoser, and CAP cover only 10, 9, and 7 tasks, respectively (\cref{table:zero-shot}). } GeneralVLA outperforms the baselines in 10 out of the 14 tasks. The three lowest-performing tasks for GeneralVLA are non-prehensile or complex fine-grained manipulation tasks that require more precise visual 3D pose estimation or
dynamic adjustment during execution. VoxPoser fails in tasks that require moving the arm beyond 4-DoF. 
We further tested the multi-stage execution capability (\cref{fig:examplerollout}). The results show that GeneralVLA can generalize to multi-object, multi-stage tasks.

\begin{table*}[t]
\caption{\textbf{Behavior Cloning with different generated data.} The behavior cloning policy trained on the data generated by GeneralVLA provides the best performance on 10 out of 12 tasks compared to the other autonomous data generation baselines (excluding RLBench). We report the Success Rate \% for behaviour cloning policies trained with data generated from VoxPoser~\cite{DBLP:conf/corl/HuangWZL0023} and Code as Policies~\cite{DBLP:conf/icra/LiangHXXHIFZ23} in comparison. Note that the RLBench~\cite{DBLP:journals/ral/JamesMAD20} baseline uses human expert demonstrations and is considered an upper bound for behavior cloning.}
\centering
\vspace{-5pt} 
    \resizebox{1.\textwidth}{!}{
    \begin{tabular}{cccc cccc} %
      \toprule  %
      \enspace \textbf{Data}   & \textbf{Models}  & Put\_block & Play\_jenga & Open\_jar &Close\_box & Open\_box & Pickup\_cup  \\
      \midrule
      VoxPoser~\cite{DBLP:conf/corl/HuangWZL0023} & RVT-2~\cite{DBLP:conf/rss/0001B0GCF24} &  2.67±2.31 & -  & - & - & - & 4.00±4.00
\\
      CAP~\cite{DBLP:conf/icra/LiangHXXHIFZ23} & RVT-2~\cite{DBLP:conf/rss/0001B0GCF24} & 6.67±2.31 & - & - & - & - &  14.67±12.86
\\
      Scaling-up~\cite{DBLP:conf/corl/HaFS23} & RVT-2~\cite{DBLP:conf/rss/0001B0GCF24} & 22.67±15.14  & - & 5.33±6.11 & - & - & 14.67±2.31
\\
      GeneralVLA (Ours) & RVT-2~\cite{DBLP:conf/rss/0001B0GCF24} & \textbf{86.67}±3.06  & 82.67±9.45 & \textbf{21.67}±13.05 & 54.00±12.00 & \textbf{32.67}±14.19 & 56.00±5.29
\\
      RLBench~\cite{DBLP:journals/ral/JamesMAD20} & RVT-2~\cite{DBLP:conf/rss/0001B0GCF24} & 20.00±18.33  & \textbf{81.33}±9.24 & \textbf{58.67}±45.49 & \textbf{68.00}±24.98 & 14.67±6.11 & \textbf{54.67}±23.09
\\
      \bottomrule %
      \toprule
        \enspace \textbf{Data}   & \textbf{Models}   & Take\_umbrella & Sort\_mustard & Open\_wine & Lamp\_on & Put\_knife & Pick\_\&\_lift \\
      \midrule
      VoxPoser~\cite{DBLP:conf/corl/HuangWZL0023} & RVT-2~\cite{DBLP:conf/rss/0001B0GCF24} &  4.00±4.00 & 0.00±0.00  & 1.33±2.31 & 5.33±4.62 &  1.33±2.31 & 5.67±1.64
\\
      CAP~\cite{DBLP:conf/icra/LiangHXXHIFZ23} & RVT-2~\cite{DBLP:conf/rss/0001B0GCF24} & 13.33±10.06 & - & - & 8.00±16.00 & 9.33±6.11 & 46.67±2.31
\\
      Scaling-up~\cite{DBLP:conf/corl/HaFS23} & RVT-2~\cite{DBLP:conf/rss/0001B0GCF24} & 4.00±4.00  & 0.00±0.00 & 81.33±12.86 & 76.00±4.00 & 5.33±2.31 & 53.33±10.06
\\
      GeneralVLA (Ours) & RVT-2~\cite{DBLP:conf/rss/0001B0GCF24} & \textbf{87.33}±5.03 & 60.67±7.57 & 81.33±8.33 & \textbf{88.67}±4.16 & 8.00±4.00 & 47.33±4.16 
\\
      RLBench~\cite{DBLP:journals/ral/JamesMAD20} & RVT-2~\cite{DBLP:conf/rss/0001B0GCF24} & 58.67±50.80  & \textbf{53.33}±34.02& \textbf{86.67}±12.86 & 84.00±13.86 & \textbf{30.67}±10.07 & \textbf{62.67}±9.24
\\
      \bottomrule %
    \end{tabular}
    }
    \label{table:bc}
    \vspace{-10pt} 
\end{table*}

\subsection{Behavior cloning with demonstrations from GeneralVLA}

GeneralVLA are computationally expensive with the present of ASM and 3DAgent. Besides, it fail to fully leverage the experience from previous successful plans due to the replanning from scratch for each time.
It holds the potential to generate useful training data. 
Therefore, we record the trajectories generated by zero-shot in~\cref{Zero-shot Performance in Simulation} and distill them into a policy network to improve the execution efficiency of the robotic arm. This also fully utilizes the experience from successful plans to further reduce the error rate.
We also compare performance against a model trained on human-generated demonstrations across the 12 tasks. 
We use the data to train behavior cloning policies.

\begin{minipage}[t]{0.28\textwidth}
\centering
\captionof{table}{\textbf{Quantitative comparisons on object reference (RoboRefIt).} The metric is percentage of predicted points within the target mask.}
\vspace{-5pt} 
\small 
\resizebox{\textwidth}{!}{ %
\begin{tabular}{ccc}
\toprule
\enspace \textbf{Method}   & \textbf{Reference}  & \textbf{Accuracy}    \\
\midrule
\textbf{GPT-4o~\cite{gpt-4o}} & OpenAI 24 & 15.3±1.3 \\
\midrule
\textbf{LLaVA-NeXT~\cite{liu2024llava}} & LLaVA 24 & 20.0±0.9 \\
\midrule
\textbf{Qwen2.5-VL~\cite{bai2025qwen2}} & Qwen 25 & 24.1±0.9 \\
\midrule
\textbf{SpatialLLM~\cite{ma2025spatialllm}} & CVPR 25 & 21.3±0.9 \\
\midrule
\textbf{ASM} & Ours & 63.4±1.4 \\
\bottomrule
\end{tabular}
}
\label{table:pointprediction}
\end{minipage}%
\hfill %
\begin{minipage}[t]{0.165\textwidth}
\centering
\captionof{table}{\textbf{Ablation on the data composition.} }
\vspace{-5pt} 
\resizebox{\textwidth}{!}{
\begin{tabular}{cc}
\toprule
\enspace \textbf{Data}  & \textbf{Accuracy}   \\
\midrule
{No VQA}&52.1±2.8 \\
\midrule
{No LVIS}&32.1±2.2 \\
\midrule
{No Pixel}&48.2±1.1 \\
\midrule
{No Sim}&51.9±1.7 \\
\midrule
{No Robo}&56.2±1.8 \\
\midrule
{All}&63.4±1.4 \\
\bottomrule
\end{tabular}
}
\label{table:datacomposition}
\vspace{0.4cm} 
\end{minipage}

\noindent \textbf{Data generation details.} We generate 10 successful demonstrations per task. We use the system’s success condition to filter for successful demonstrations. Each of the demonstrations consists of a language instruction, RGB-D frames for the trajectory, and waypoints represented as 6 DoF gripper poses and states. For the tasks where the baselines were unable to generate any successful demonstrations, we patched the missing training data with RLBench system-generated demonstrations.

\noindent \textbf{Training and evaluation protocol. } 
We train a model using the generated demonstrations: the RVT-2 model~\cite{DBLP:conf/rss/0001B0GCF24}, a transformer-based robotic manipulation behavior cloning model that expects tokenized voxel grids and language instructions as inputs and predicts discretized voxel grid 6 DoF poses and gripper states.
For all the generated training datasets, we train a multi-task RVT-2 policy with a batch size of 4 for 30k iterations on a single RTX A40. To ensure consistent evaluation, we generate one set of testing environments with RLBench. We evaluate the last checkpoint from each of the trained policies. Each policy is evaluated for 50 episodes across each task using 3 different seeds. We measure the success rate based on the simulation-defined success condition.

\noindent \textbf{Results:} Policies trained using GeneralVLA data perform similarly to policies trained using hand-scripted demonstrations for RVT-2 (\cref{table:bc}). 
Training on either GeneralVLA or hand-scripted demonstrations results in a performance difference of just 2.7\% on average across all tasks. 
Furthermore, models trained on data from the baselines exhibit statistically lower performance (VoxPoser, Scaling-up, and CAP). One of the main factors potentially contributing to the performance differences could be that GeneralVLA generates diverse expert trajectories that are preferable to humans. 
We also observe that the policy trained on GeneralVLA data achieves a lower standard deviation of 6.24, on average, across all tasks, compared to the zero-shot performance standard deviation of 11.02. This suggests the benefits of training on generated data instead of relying solely on zero-shot deployment.

\begin{figure}
  \centering
  \includegraphics[width=.5\textwidth]{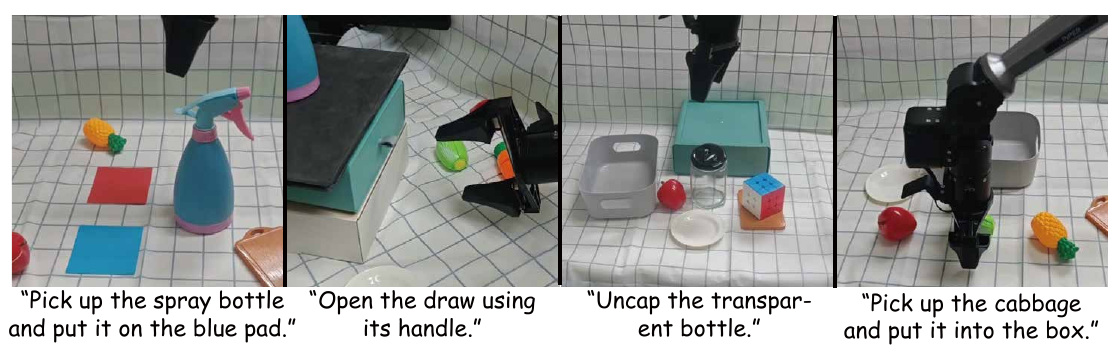}
  \vspace{-20pt} 
  \caption{GeneralVLA is an open-vocabulary robot demonstration generation system. We show zero-shot demonstrations for 4 tasks in the real world.} %
  \label{fig:realtask}
    \vspace{-0.4cm} 
\end{figure}

\subsection{Real-world experiments}

\textbf{Environment and tasks.} We conduct zero-shot tests on an Agilex-2.0 Piper manipulator equipped with a parallel gripper. We use a top-facing Intel RealSense L515 LiDAR RGB-D camera. We selected 4 representative real-world tasks: move\_spray\_bottle,  open\_drawer, open\_jar, sort\_object, all conditioned on language instructions. Each task was evaluated over 10 episodes with varying object poses across 3 trials.

\textbf{Results:} GeneralVLA is able to generate successful demonstrations for each of the 4 real world tasks. 
Taking Move\_spray\_bottle as an example, GeneralVLA can calculate the height of the bottle using 3DLLM and provide a suitable placement position, while Robopoint lacks trajectory planning. In the Open\_draw task, 3DLLM can determine the orientation of the drawer, whereas CAP does not have pre-designed hand-crafted primitive actions for opening drawers.

\section{Ablation Study}

\subsection{Point Location Accuracy of ASM} %

\begin{table}[t]
\setlength{\tabcolsep}{1.5pt}
\caption{We present a comparison of success rates for task completion in a zero-shot manner (Code as Policies~\cite{DBLP:conf/icra/LiangHXXHIFZ23} and GeneralVLA)} %
\centering
\vspace{-5pt} 
    \resizebox{.48\textwidth}{!}{
    \begin{tabular}{ccccc} %
      \toprule  %
      \enspace \textbf{Method}   & Move\_spray\_bottle &  Open\_drawer & Open\_jar & Sort\_object     \\
      \midrule
      CAP (0-shot) &  6.67  & 0.00 & 36.67  & 70.00
\\
      Robopoint (0-shot) &  0.00  & 0.00 &  20.00  & 63.33 
\\
      GeneralVLA (0-shot) &  \textbf{63.33.00 } & \textbf{36.67} & \textbf{ 50.00}  & \textbf{76.67 }
\\
      \bottomrule %
    \end{tabular}
    }
    \label{table:realworld}
    \vspace{-0.4cm} 
\end{table}

In \cref{table:pointprediction}, we report the average prediction accuracy for ASM and the baselines, along with the standard deviation computed from 3 different runs. The accuracy is calculated as the percentage of predicted points within the ground truth target mask. We can see that ASM achieves significantly higher accuracy than all baselines, demonstrating the power of ASM in spatial reasoning and precise target generation.

In~\cref{table:datacomposition}, we evaluated the importance of each data component on the RoboRefIt benchmark. Each data component – VQA on real images, object detection from LVIS, object reference on real and synthetic images – significantly contributes to overall accuracy. This highlights the value of a general problem formulation that incorporates diverse data sources.
Among them, LVIS provides precise semantic information and location information, playing an important role.

\subsection{Information Required by 3DAgent} %

In the 3DAgent planning process, we designed multiple components, including 3D point information input, with each object having no fewer than 3 points and incorporating target object and obstacle information. We conducted ablation experiments on these components (\cref{table:3DAgent}).
Among them, 3DAgent-2D performs trajectory reasoning using only 2D point information and projects the inferred trajectory into 3D space for execution after reasoning. 3DAgent-1point includes only 1 3D point per object. In this case, 3DAgent cannot determine the orientation of objects or obstacles through multiple points. 3DAgent w/o obstacle only recognizes the position of the operated object, failing to achieve obstacle avoidance or determine obstacle orientation. For example, it cannot judge the orientation of an umbrella bag, thus failing to pull out the umbrella in the correct direction.
The results show that the settings of 3D point information input, with no fewer than 3 points per object, and the inclusion of both target object and obstacle information are necessary, improving the success rate of trajectories planned by 3DAgent.

\begin{figure}
  \centering
  \includegraphics[width=.5\textwidth]{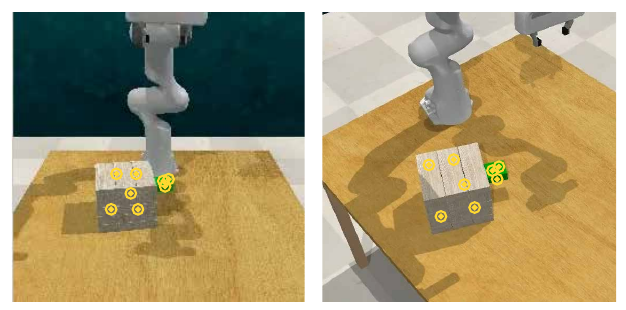}
  \vspace{-20pt} 
  \caption{\textbf{The multi-view robustness of ASM.} This assist 3DAgent in reasoning about the direction for pulling out the block. } %
  \label{fig:viewpoints}
    \vspace{-0.2cm} 
\end{figure}

\begin{table}[t] \small
\caption{Ablation Study on 3DAgent for Trajectory Planning.} %
\centering
\vspace{-5pt} 
    \resizebox{.8\linewidth}{!}{
    \begin{tabular}{ccc} %
      \toprule  %
      \enspace \textbf{Method}   & Take umbrella  & Put block     \\
      \midrule
      3DAgent-2D &  4.00±4.00  & 19.33±3.06 
\\
      3DAgent-1point &  24.00±4.00  & 82.00±7.21 
\\
      3DAgent w/o obstacle &  23.33±7.57  & 91.33±4.16 
\\
      3DAgent & 67.33±14.05  & 93.33±3.06
\\
      \bottomrule %
    \end{tabular}
    }
    \label{table:3DAgent}
    \vspace{-0.4cm} 
\end{table}

\begin{table}[t]
\caption{Ablation Study on the HGM Module.} %
\centering
\vspace{-5pt} 
    \resizebox{.8\linewidth}{!}{
    \begin{tabular}{ccc} %
      \toprule  %
      \enspace \textbf{Method}   & Play jenga & Take umbrella   \\
      \midrule
      HGM w/o rgb &  56.67±4.53  & 32.33±14.03 
\\
      HGM w/o 3D point &  0.00±0.00  & 0.00±0.00 
\\
      HGM w/o filter-C &  58.00±5.52  & 53.00±12.41 
\\
      HGM w/o filter-N &  76.33±7.24  & 54.67±14.00 
\\
      HGM &  84.67±11.02 & 67.33±14.05
\\
      \bottomrule %
    \end{tabular}
    }
    \label{table:HGM}
    \vspace{-0.2cm} 
\end{table}

\subsection{Ablation Study of the HGM Module} %
The HGM module integrates multi-modal input information, including RGB images, depth maps, and 3D point ranges. It also designs a collision filter-out mechanism and a nearest-selection mechanism. The specific process is further organized in the appendix. We conducted ablation experiments on the above designs (\cref{table:HGM}).
Among them, HGM w/o rgb does not include RGB information and only uses depth maps to estimate grasping poses. HGM w/o 3D point does not use the 3D points and their semantic information estimated by ASM, making HGM unable to determine the category of the grasped object. HGM w/o filter-C does not perform collision detection on multiple grasping poses. In this case, the estimated grasping pose may collide with surrounding obstacles or embed into the object, affecting grasping accuracy. HGM w/o filter-N does not select the grasping pose closest to the object’s center point but randomly selects grasping poses that meet other conditions.
The results indicate that multi-modal information from RGB images, depth maps, and 3D point ranges is necessary for task completion, and collision detection as well as optimal pose selection can also improve the algorithm’s success rate.

\begin{figure}
  \centering
  \includegraphics[width=.4\textwidth]{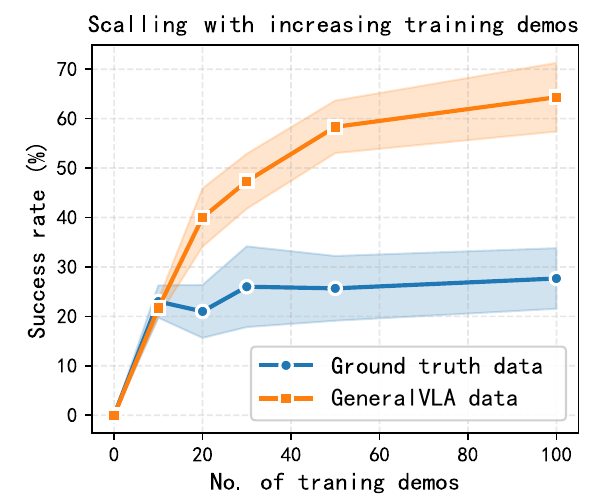}
  \vspace{-10pt} 
  \caption{\textbf{Scaling experiment.} Scaling effect of model performance with increasing training demonstrations. } %
  \label{fig:Scaling}
    \vspace{-0.4cm} 
\end{figure}

\subsection{Date scaling}
For the effective further development of GeneralVLA, it is crucial that the collected data supports the scaling of robotics tasks. We conducted an ablation study to evaluate the quality of the GeneralVLA-generated data for scaling.
Our scaling experiments demonstrate that generating more training data via GeneralVLA improves RVT-2 policy performance (\cref{fig:Scaling}). The data from our approach show a better rate of change with a slope of 0.539 for a linear fit, compared to 0.178 for RLBench-generated data.

\section{Conclusion}

In this work, we introduced GeneralVLA, a hierarchical vision-language-action framework that enables zero-shot robotic manipulation and high-quality data generation without requiring real-world robotic data.
Our key innovation lies in the three-tier architecture that effectively leverages foundation models' world knowledge: ASM for precise affordance perception, 3DAgent for spatial reasoning and trajectory planning, and HGM for fine-grained manipulation.
It performs well in scenarios requiring 3D reasoning and also features obstacle avoidance capabilities.

Given that current VLMs exhibit limitations in precise spatial pose estimation, our work utilizes them only for 2D point estimation.
Future work could enhance VLM's spatial perception capability by incorporating 3D pose estimation.
We believe GeneralVLA establishes a promising direction for leveraging foundation models in robotics, demonstrating that hierarchical decomposition combined with world knowledge can overcome data scarcity challenges while achieving robust generalization.

\bibliographystyle{plainnat}
\bibliography{references}

\clearpage
\section*{Appendix}

\section{VLM Finetuning Dataset Details}
Our dataset for fine-tuning consists of 5 different sources. %

\noindent \textbf{Pixel Point Pred Data.} Our point prediction dataset comes from Robopoint~\cite{DBLP:conf/corl/YuanDBPKMMF24}. 347k samples in our point prediction dataset contain labels given as a set of unordered points, such as $p^o = [(0.25, 0.11), (0.22, 0.19), (0.53, 0.23)]$, or bounding boxes in $[(cx, cy, w, h)]$ style.
Most answers are represented as a list of 2D points corresponding to locations on the image.

\noindent \textbf{LVIS} is constructed from~\cite{DBLP:conf/cvpr/GuptaDG19}. Unlike~\cite{DBLP:conf/corl/YuanDBPKMMF24}, which focuses on bounding boxes, we pay more attention to the semantic regions of objects. We randomly sample 2D points from segmentation masks and attach corresponding semantic information to them.
This teaches the model how to ground language to image regions.

\noindent \textbf{Robot Data.} Using robot data allows us to ensure that the VLM can reason about objects and robot gripper positions. 
We use existing online robot datasets that are not from the deployment environment to enable this VLM capability.
We source 100k points from the Open X-Embodiment dataset~\cite{DBLP:conf/icra/ONeillRMGPLPGMJ24}, consisting of a Jaco Play arm (a different embodiment from the test robot) performing manipulation tasks.
We additionally generate a dataset of simulated robotics tasks from SIMPLER~\cite{DBLP:conf/corl/LiHGMPWFLSKL0F024}. The point positions and categories are resolved via proprioception, camera parameters, and task descriptions. %

\noindent \textbf{VQA data.} The VQA data is a mix of 667K conversations from~\cite{DBLP:conf/iccv/KafleK17} where the model is asked to answer questions in natural language based on the input image, such as “What is the person feeding the cat?”, We keep this data as is because these VQA queries are likely to benefit a VLM’s semantic reasoning and visual generalization capabilities This ensures the model can reason in language. 
We formulate different data sources into the same format and co-train with all of them.

\begin{figure*}
  \centering
  \includegraphics[width=0.8\linewidth]{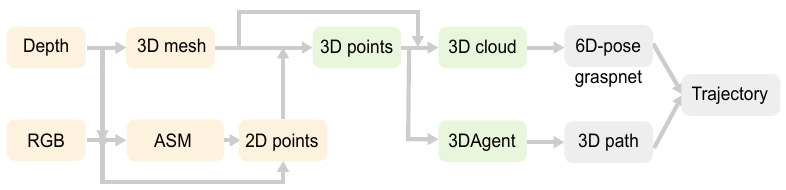}
  \caption{\textbf{Dataflow diagram of GeneralVLA.} The inputs of GeneralVLA are a depth map and an RGB image, while the output is the motion trajectory of the robotic arm’s end-effector.} %
  \label{fig:dataflow}
    \vspace{-10pt} 
\end{figure*}

\section{Implementation and Architecture Details}
\label{Implementation and Architecture Details}

\noindent \textbf{Dataflow of the GeneralVLA algorithm.} %
To provide a more comprehensive illustration of the algorithm’s execution details, we present the dataflow diagram in~\cref{fig:dataflow}. 
The algorithm takes a depth map and an RGB image as input and produces the final end-effector motion trajectory of the robotic arm.
Specifically, the RGB image is processed by ASM to detect task-relevant objects and generate their 2D pixel coordinates (2D points).
The depth map and RGB image are fused to construct a full 3D mesh of the scene.
The 2D points are projected into 3D space using the camera intrinsic parameters, yielding the corresponding 3D coordinates of the objects (3D points).
These 3D points are then used in two parallel branches. On the one hand, the 3D points are fed to 3DAgent, which, according to the task description, plans a sequence of waypoints required to complete the task, resulting in a 3D path.
On the other hand, to grasp the task-relevant object, the corresponding 3D points are used to crop the full 3D mesh, producing a localized point cloud (3D cloud) that contains only the points in the vicinity of the target object. This 3D cloud is then passed to graspnet for grasp pose estimation. The candidate grasps are further filtered using the selection rules implemented in HGM, yielding the optimal 6D grasp pose (6Dpose).
Finally, the 6D pose and the 3D path are combined to form the complete motion trajectory of the robotic arm’s end-effector.
The entire pipeline operates in a zero-shot manner and generates trajectories with high success rates.

\noindent \textbf{Spatial Affordance Prediction.} We formulate the problem of spatial affordance prediction as predicting a set of target point coordinates $\{object:(x_0, y_0), (x_1, y_1), ..., (x_n, y_n)\}$ in image space that satisfy the relations indicated by a language prompt. 
To further improve prediction accuracy, we perform multiple rounds of recognition on the initially detected points. After each detection yields a coordinate, we crop a quarter of the original image centered on that point and use this smaller region as the new input for the next recognition step. Each point is recognized three times in total.
This formulation has several advantages. First, compared to fuzzy language actions such as “place the apple in the drawer”, which requires additional detection of apple and drawer before execution, a point prediction is much more precise and places greater emphasis on affordance. Most VLMs are trained to predict bounding boxes. However,
bounding boxes often include a lot of undesirable clutter due to camera perspective and are not as specific as point outputs. Second, our formulation is general enough to enable various robotic tasks. For example, the predicted points can be interpreted as contact points for grasping, region proposals for placement or waypoints for navigation. This not only allows the model to perform multiple tasks but also means it can be trained with multi-task data. %

\noindent \textbf{Instruction Fine-tuning.}
Specifically, we follow the instruction tuning pipeline in Liu et al.~\cite{DBLP:conf/nips/LiuLWL23a}. As shown in~\cref{fig:pipline}A, the model consists of an image encoder, a MLP projector, a language tokenizer and a transformer vision language model. The image encoder processes the image into a set of tokens which are then projected by a 2-layer MLP into the same space as the language tokens. The multimodal tokens are concatenated and passed through the language transformer. All modules are initialized with pre-trained weights. The projector and the transformer weights are allowed to update while the vision encoder and tokenizer weights are frozen. The model is autoregressive and the objective is to predict the response tokens and a special token delineating the boundary between instruction and response.

\noindent \textbf{Action Generation Module.} We generate each action using object-centric approach. We utilize M2T2~\cite{DBLP:conf/corl/YuanMMF23}, NVIDIA’s foundational grasp prediction model, for pick and place actions. For 6-DoF grasping, we input a single 3D point cloud from RGB-D camera. The model outputs a set of grasp proposals on any graspable objects, providing 6-DoF grasp candidates (3-DoF rotation and 3-DoF translation) and default gripper close states. For placement actions, M2T2 outputs a set of 6-DoF placement poses, indicating where the end-effector should be before executing a drop action based on a trajectory plan. The network ensures the object is stably positioned without collisions. We also set default values for mask\_threshold and object\_threshold to control the number of proposed grasp candidates. As illustrated in the main text,  the 3D point information can locate the 3D spatial range of the object. After determining the 3D spatial range, the scope of candidate grasp poses are determined.
We  also select the grasp pose whose grasp center is closest to the object's center point.
This approach allows us to obtain the most optimal task-specific grasp pose, placement pose, and pre-action pose for manipulation tasks, all leveraging the capabilities of the foundation model. After the action is generated, a motion planner can be used to move the end-effector to the desired pose as detailed in~\cref{fig:pipline}C.

\section{Experiments Details}

\begin{figure}
  \centering
  \includegraphics[width=.5\textwidth]{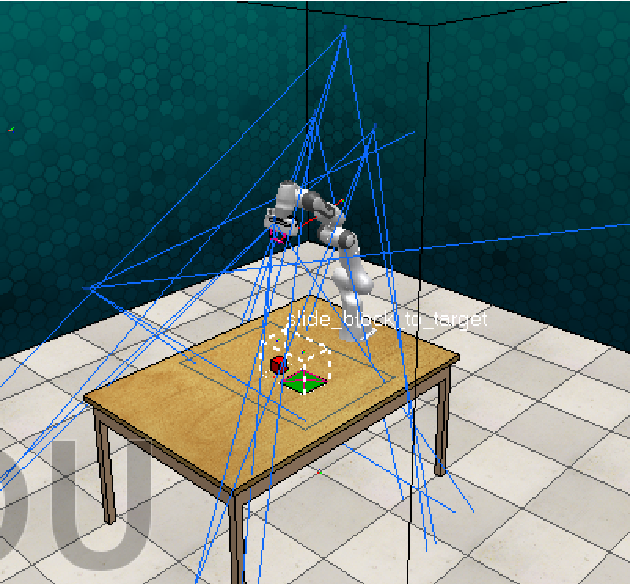}
  \vspace{-20pt} 
  \caption{Simulation scene setup.} %
  \label{fig:rlbench}
    \vspace{-10pt} 
\end{figure}

\noindent \textbf{Simulation Setup.} All the simulated experiments use a four-camera setup as illustrated in~\cref{fig:rlbench}. The cameras are positioned at the front, left shoulder, wrist, and right shoulder. All cameras are static, except for the wrist camera, which is mounted on the end effector. We did not modify the default camera poses from the original RLBench~\cite{DBLP:journals/ral/JamesMAD20}. These poses maximize coverage of the entire table, and we use a 256 x 256 resolution for better input to the VLMs.  

\noindent \textbf{Task Details.} We describe in detail each of the 12 tasks for simulation evaluation, both for trained policies and zero-shot methods, along with their RLBench variations and success conditions. We have made some modifications to the original tasks to enhance the detection rate by Code-As-Policies and VoxPoser.

\subsection{\rm put block  }

\textbf{Filename: } put\_block.py  

\noindent \textbf{Task:} Pick up the green block and place it on the red mat.  

\noindent \textbf{Success Metric:} The success condition on the red mat detects the target green block.  

\subsection{\rm close box }   

\textbf{Filename:} close\_box.py  

\noindent \textbf{Task:} Close the box.  

\noindent \textbf{Success Metric:} The revolute joint of the specified handle is at least 60◦ off from the starting position. 

\subsection{\rm open box  }   

\textbf{Filename: }open\_box.py  

\noindent \textbf{Task:} Open the box.  

\noindent \textbf{Success Metric:} The revolute joint of the specified handle is at least 60◦ off from the starting position.  

\subsection{\rm play jenga }   

\textbf{Filename:} play\_jenga.py  

\noindent \textbf{Task:} Pull out the green jenga block.  

\noindent \textbf{Success Metric:} The green jenga block is out of its pre-defined location.

\subsection{\rm  open jar   }

\textbf{Filename:} open\_jar.py  

\noindent \textbf{Task:} Uncap the green jar.  

\noindent \textbf{Success Metric:} The green jar is out of its pre-defined capped location.

\begin{figure}
  \centering
  \includegraphics[width=.5\textwidth]{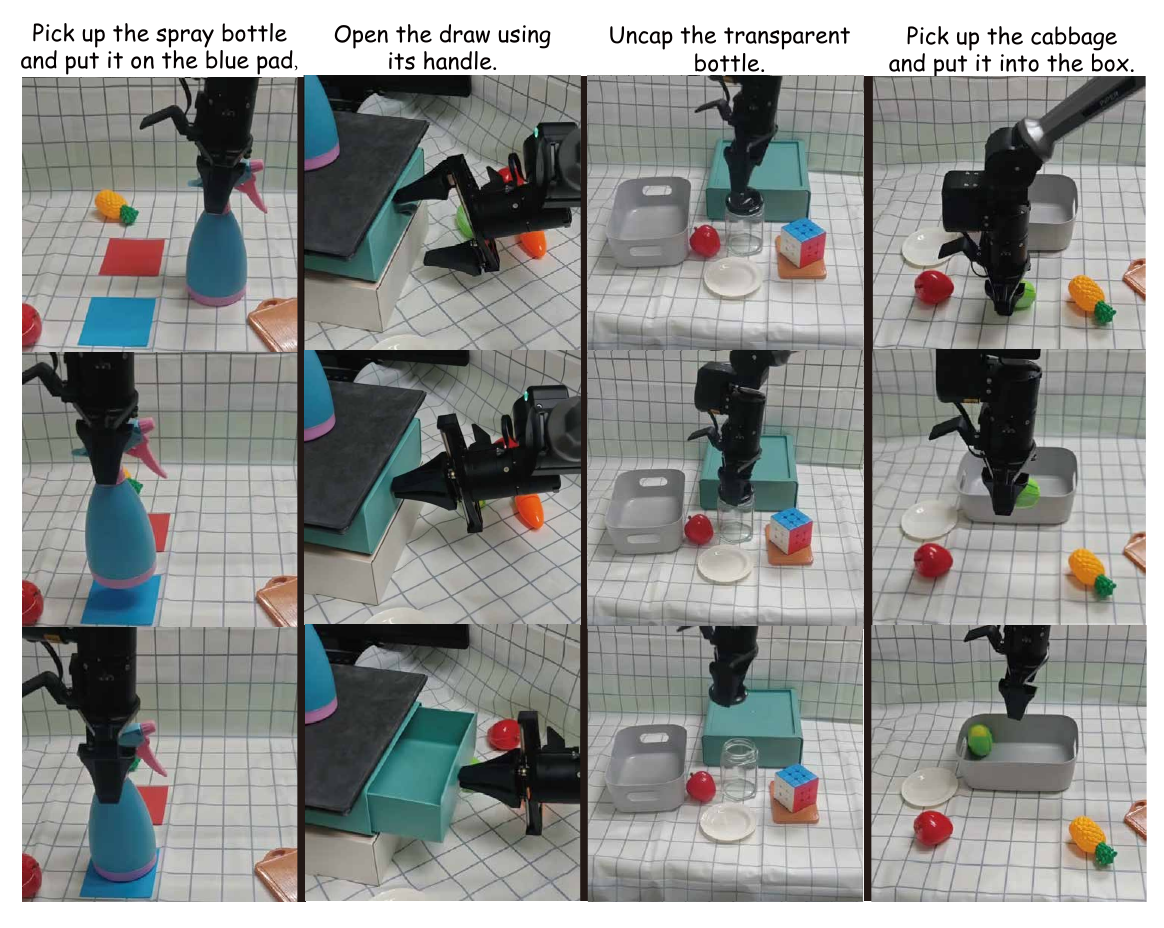}
  \caption{Task execution in four real-world scenarios, with each column representing one task. From left to right, the images illustrate the execution process of the four tasks listed in~\cref{table:realworld} } %
  \label{fig:realworld}
    \vspace{-10pt} 
\end{figure}

\subsection{\rm pickup cup  } 

\textbf{Filename:} Filename: pickup\_cup.py  

\noindent \textbf{Task:} Pick up the red cup.  

\noindent \textbf{Success Metric:} Lift up the red cup above the pre-defined location.  

\subsection{\rm take umbrella   } 

\textbf{Filename:} take\_umbrella\_out\_of\_stand.py  

\noindent \textbf{Task:} Pick up the umbrella out of the umbrella stand.  

\noindent \textbf{Success Metric:} Lift up the umbrella out of the umbrella stand. 

\subsection{\rm sort mustard   }

\textbf{Filename:} sort\_mustard.py  

\noindent \textbf{Task:} Pick up the yellow mustard bottle, and place it into the red container.

\noindent \textbf{Success Metric:} The yellow mustard bottle inside red container. 

\subsection{\rm open wine }  

\textbf{Filename:} open\_wine.py 

\noindent \textbf{Task:} Uncap the wine bottle.  

\noindent \textbf{Success Metric:} The wine bottle cap is out of its original position.  

\subsection{\rm lamp on }   

\textbf{Filename:} lamp\_on.py  

\noindent \textbf{Task:} Turn on the lamp.  

\noindent \textbf{Success Metric:} The lamp light up.  

\subsection{\rm put knife  } 

\textbf{Filename:} put\_knife\_on\_chopping\_board.py  

\noindent \textbf{Task:} Pick up the knife and place it onto the chopping board.  

\noindent \textbf{Success Metric:} Knife on chopping board.  

\subsection{\rm push block}   

\textbf{Filename:} push\_block\_to\_target.py  

\noindent \textbf{Task:} Push the red block down towards the green target.  

\noindent \textbf{Success Metric:} The red block fails within the green target. 

\subsection{\rm insert block  } 

\textbf{Filename: }insert\_block.py  

\noindent \textbf{Task:} Push the green block into the jenga tower.  

\noindent \textbf{Success Metric:} The green block inserted in.

\subsection{\rm  pick \& lift  } 

\textbf{Filename:} pick\_and\_lift.py  

\noindent \textbf{Task:} Pick up the red cube.  

\noindent \textbf{Success Metric:} The red cube is lifted up.

\section{Prompts}

\textbf{ASM Prompt.} We list the prompt for zero-shot manipulation and evaluation in~\cref{fig:pipline}. We condition the model on an image and the prompt, except when training on affordance prediction data where we used the given prompts from the dataset. We ask the model to output the position of each object related with task. Each object will contains more than 3 points to describe its position and spacial pose.

\noindent \textbf{3DAgent Prompt.} We list the prompt for zero-shot manipulation and evaluation in~\cref{fig:pipline}. We condition the model on the prompt, which contains the 3D position with the name of related objects. 
Note that we ask the model to output gripper changes as separate language tokens, i.e., Open Gripper/Close Gripper, as opposed to as a dot as shown in simplified depictions like~\cref{fig:examplerollout}.

\begin{tcolorbox}

\textbf{ASM Prompt}

In the image, please describe the related object in task described in ⟨quest⟩{quest}⟨/quest⟩.

Provide a list of points denoting the affordance position of related objects. 

Format your answer as a list of tuples enclosed by ⟨ans⟩ and ⟨/ans⟩ tags. For example: 
$⟨ans⟩[[cube,(0.25, 0.21),(0.22, 0.23),(0.23, 0.24)], \\
...]⟨/ans⟩  $

The tuple denotes the x, y location of the object in the image.
Each object contains more than 3 points.

The coordinates should be floats ranging between 0 and 1, indicating the relative locations of the points in the image, with (0,0) at the bottom-left corner.

\end{tcolorbox}

\begin{tcolorbox}

\textbf{3DAgent Prompt}

Please execute the command described in ⟨quest⟩{quest}⟨/quest⟩. 

The coordinates of objects in the scene are
$⟨ans⟩[[cube,(0.25, 0.21,0.11),(0.22, 0.23,0.10),\\ (0.23, 0.24, ,0.11)], 
...]⟨/ans⟩  $

Provide a sequence of points denoting the trajectory of a robot gripper to achieve the goal. 

Format your answer as a list of tuples enclosed by ⟨ans⟩ and ⟨/ans⟩ tags. For example: 

$⟨ans⟩[(0.25, 0.32, 0.10), (0.32, 0.17, 0.10), \\ ⟨action⟩  Close Gripper⟨/action⟩, (0.13, 0.24, \\ 0.10),  ⟨action⟩Open Gripper⟨/action⟩,(0.74,\\ 0.21, 0.20), ⟨action⟩Grasp⟨/action⟩,   ...]⟨/ans⟩  $

The tuple denotes the x, y and z location of the end effector of the gripper in the space. The action tags indicate the gripper action.

The coordinates should be floats ranging between 0 and 1, indicating the relative locations of the points in the space.
The points on the trajectory should not exceed 20.

\end{tcolorbox}

\section{Failure Analysis}

\subsection{Different Failure Modes}

\noindent \textbf{Affordance Prediction Failures} 

\noindent - Failure to understand the scene: %
ASM may err in task understanding, failing to extract task-relevant object information or lacking sufficient visual-semantic perception to provide correct object categories.

\noindent - Incorrect trajectory prediction: The ASM may fail to predict the precise 2D points of target objects. 
For example, mislabeling a box’s position onto the tabletop can lead to incorrect 3D pose estimation, or failing to mark a small block due to its size.

\noindent \textbf{Trajectory Prediction Failures}

\noindent - Failure to understand the language goal: Although the 3DAgent demonstrates strong capabilities in handling diverse text question, it can misunderstand the goal and make inaccurate predictions.

\noindent - 3D ambiguity: 
Insufficient 3D points may prevent accurate box pose estimation, or limited reasoning may fail to infer the 3D direction for extracting a block.

\noindent - Incorrect trajectory prediction: %
3DAgent may fail to avoid obstacles, resulting in planned trajectories that collide with objects.

\noindent \textbf{Action Execution Failures} 

\noindent - Incorrect object interaction: The low-level action model is not explicitly constrained to strictly follow the predicted trajectory. As a result, it may deviate, interacting with the wrong object and causing task failures.

\noindent - Trajectory Adherence Failures 

The policy may fail during execution. For example, in grasping tasks, an incorrect grasp angle can cause the object to slip, resulting in a failed grasp.

\begin{figure}
  \centering
  \includegraphics[width=.4\textwidth]{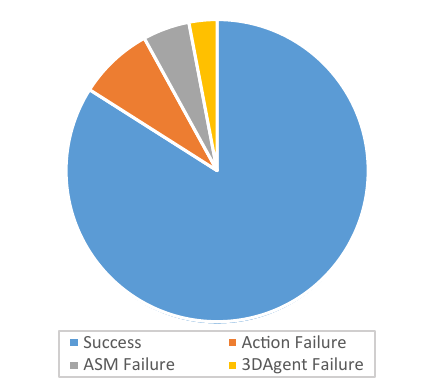}
  \caption{Performance Distribution of GeneralVLA in play jenga task.} %
  \label{fig:Errorbreakdown}
    \vspace{-10pt} 
\end{figure}

\subsection{Failure Analysis}

Our analysis in~\cref{fig:Errorbreakdown} reveals distinct failure tendencies in space reasoning and manipulation task.

In the Play Jenga task, GeneralVLA successfully completes the task in 82\% of cases. This demonstrates that ASM accurately identifies object positions, 3DAgent correctly infers the box’s 3D pose, and HGM successfully executes grasping.
However, most failures occur during the action execution phase. Precise object manipulation remains a challenging problem; despite our carefully designed grasping module, grasping failures are still possible.
ASM exhibits a low failure rate, thanks to our targeted improvements in its object position recognition capability.
3DAgent, leveraging its strong semantic reasoning ability, produces few failure cases.

\end{document}